\documentclass[conference]{IEEEtran}
\IEEEoverridecommandlockouts
\usepackage{cite}
\usepackage{amsmath,amssymb,amsfonts}
\usepackage{algorithmic}
\usepackage{graphicx}
\usepackage{textcomp}
\usepackage{xcolor}
\usepackage{multirow}
\usepackage{jabbrv}

\usepackage[super]{nth}
\def\BibTeX{{\rm B\kern-.05em{\sc i\kern-.025em b}\kern-.08em
    T\kern-.1667em\lower.7ex\hbox{E}\kern-.125emX}}

\makeatletter 
\newcommand{\linebreakand}{%
  \end{@IEEEauthorhalign}
  \hfill\mbox{}\par
  \mbox{}\hfill\begin{@IEEEauthorhalign}
}
\makeatother 

\begin{document}

\title{Neural Speech Embeddings for Speech Synthesis Based on Deep Generative Networks
\footnote{{\thanks{This work was supported by Institute for Information \& Communications Technology Planning \& Evaluation (IITP) grant funded by the Korea government (MSIT) (No.2021-0-02068, Artificial Intelligence Innovation Hub; No. 2019-0-00079, Artificial Intelligence Graduate School Program(Korea University)).}
}}
}

\author{

\IEEEauthorblockN{Seo-Hyun Lee}
\IEEEauthorblockA{\textit{Dept. of Brain and Cognitive Engineering} \\
\textit{Korea University} \\
Seoul, Republic of Korea \\
seohyunlee@korea.ac.kr}  \\

\and

\IEEEauthorblockN{Young-Eun Lee}
\IEEEauthorblockA{\textit{Dept. of Brain and Cognitive Engineering} \\
\textit{Korea University} \\
Seoul, Republic of Korea \\
ye\_lee@korea.ac.kr} \\

\and

\IEEEauthorblockN{Soowon Kim}
\IEEEauthorblockA{\textit{Dept. of Artificial Intelligence} \\
\textit{Korea University} \\
Seoul, Republic of Korea \\
soowon\_kim@korea.ac.kr} \\

\linebreakand 

\IEEEauthorblockN{Byung-Kwan Ko}
\IEEEauthorblockA{\textit{Dept. of Artificial Intelligence} \\
\textit{Korea University} \\
Seoul, Republic of Korea \\
leaderbk525@korea.ac.kr} \\

\and

\IEEEauthorblockN{Jun-Young Kim}
\IEEEauthorblockA{\textit{Dept. of Artificial Intelligence} \\
\textit{Korea University} \\
Seoul, Republic of Korea \\
j\_y\_kim@korea.ac.kr} \\

 \and 

 \IEEEauthorblockN{Seong-Whan Lee}
 \IEEEauthorblockA{\textit{Dept. of Artificial Intelligence} \\
 \textit{Korea University} \\
 Seoul, Republic of Korea \\
 sw.lee@korea.ac.kr}
 }


\maketitle
\begin{abstract}
Brain-to-speech technology represents a fusion of interdisciplinary applications encompassing fields of artificial intelligence, brain-computer interfaces, and speech synthesis. Neural representation learning based intention decoding and speech synthesis directly connects the neural activity to the means of human linguistic communication, which may greatly enhance the naturalness of communication. With the current discoveries on representation learning and the development of the speech synthesis technologies, direct translation of brain signals into speech has shown great promise. Especially, the processed input features and neural speech embeddings which are given to the neural network play a significant role in the overall performance when using deep generative models for speech generation from brain signals. In this paper, we introduce the current brain-to-speech technology with the possibility of speech synthesis from brain signals, which may ultimately facilitate innovation in non-verbal communication. Also, we perform comprehensive analysis on the neural features and neural speech embeddings underlying the neurophysiological activation while performing speech, which may play a significant role in the speech synthesis works.
\end{abstract}

\begin{small}
\end{small}

\section{INTRODUCTION}

\begin{figure*}[t]
\centering
    \includegraphics[width=0.8\textwidth]{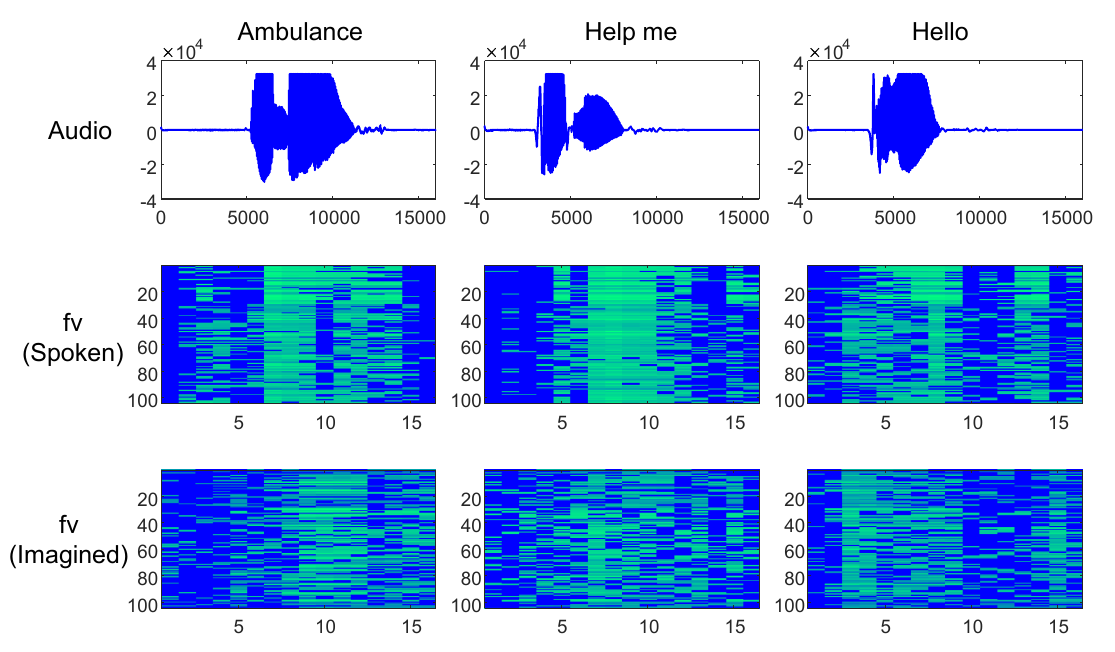}
    \caption{Feature embedding for the spoken speech and the imagined speech is demonstrated. The feature matrix was constructed using the time-wise computation of the CSP pattern, divided into 16 time points per each EEG segment. The size of the pattern for each time point was set to 104 since 8 patterns per class were computed using the multi-CSP algorithm. The value under the mean of each column was ignored to display the temporal variations of the embedding features.}
    \label{embeding}
\end{figure*}

Recently, there has been a growing interest in the field of brain-computer interfaces (BCIs) which offer a way for humans to interact with external devices or control their surroundings by using brain signals ~\cite{chaudhary2016brain}. Among the various methods used for BCI research, electroencephalography (EEG) involves recording electrical activity by placing electrodes on the scalp without the need for invasive procedures like implanting electrodes. This makes EEG a valuable source of information for applications involving brain signals ~\cite{kim2019subject, thung2018conversion}. BCIs based on EEG have been explored for a wide range of applications ~\cite{kim2023diff}, including the control of motor functions, communication, and cognitive assessment ~\cite{mane2021fbcnet, lee2019possible, lee2022toward}. Despite the challenge of low signal quality in non-invasive EEG recordings, researchers have explored numerous applications due to the ease of use and practical advantages offered by EEG ~\cite{lee2019eeg}.

Brain-to-speech (BTS) is a new stream of intuitive BCI communication which aims to generate audible speech from human brain signals \cite{lee2023AAAI}. It provides non-verbal communication facilitated by current domain adaptation and speech synthesis technologies. Neural patterns are transformed into spoken language by directly associating the speech-related features with human language. In previous studies, the domain adaptation framework established a natural correspondence between the neural features and the speech ground truth \cite{lee2023AAAI, lee2022eeg}. Therefore, audible speech could be generated from brain signals of silently imagined speech, demonstrating the potential of brain signal-mediated communication. The preprocessing procedures as well as the feature embeddings given as an input to the model is known to play a significant role in the BTS performance.

In this paper, we provide an extensive analysis of non-invasive BTS technology, an innovative field that holds promise for utilizing brain signals to synthesize speech directly. We provide an extensive analysis on the previous BTS work \cite{lee2023AAAI}, which could contribute in highly improving the overall BTS performance. This advancement could introduce a new era of non-verbal communication, transforming how we interact with the external world via brain signals. Additionally, we conducted an extensive and comprehensive examination of the neural characteristics and underlying the neurophysiological processes implicated in the act of speaking. Our research aims to clarify the intricate mechanisms involved in speech production and provide advancement in the field of neural speech synthesis. The comprehension of these neural features would enhance the understanding of human communication and provides crucial insights for advancing BTS technology.

\begin{figure*}[t]
\centering
    \includegraphics[]{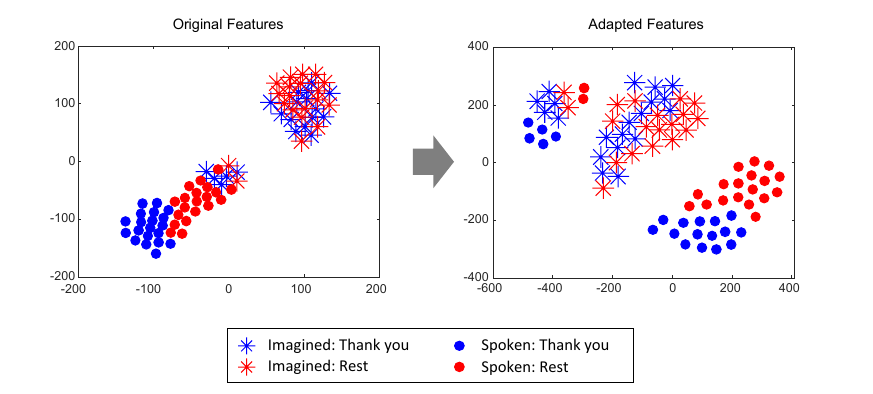}
    \caption{t-SNE plot of features before and after the adaptation process. Clusters of the imagined speech and the spoken speech have shown clear distance in the original features. However, adapted features show relatively distributed aspects across same classes in different domain (blue and red samples show broad clusters.)}
    \label{tsnel}
\end{figure*}

\section{MATERIALS AND METHODS}

\subsection{Dataset}
This research involved the use of EEG signals and voice recordings from six participants as they engaged in speaking tasks. The recordings were obtained by placing a 64-channel electrode array on the subjects' scalps to record EEG data, and a microphone was used to capture their voices. The microphone was synchronized with the EEG signals. The participants were directed to speak in accordance with instructions displayed on a screen. This experimental setup was consistent with a previous study by \cite{lee2020neural, lee2023AAAI}.


Participants were instructed to perform imagined speech and spoken speech following the instructions displayed on the screen. For the spoken speech session, the voice of each participant was recorded via a microphone in alignment with the EEG of spoken speech.

\subsection{Preprocessing}
EEG signals were segmented into 2-second intervals for each trial. The EEG data underwent filtering, which included a 5th-order Butterworth bandpass filter within the high-gamma frequency range of 30–120 Hz, a well-known frequency band associated with speech-related information according to Lachaux et al.~\cite{lachaux2012high}. Additionally, a notch filter was applied to mitigate line noise at 60 Hz and its harmonics of 120 Hz. Independent component analysis, with references from electrooculography and electromyography, was employed to remove artifacts resulting from eye and muscle activity during spoken speech. Baseline correction was carried out by subtracting the average signal from 500 ms preceding the beginning of each trial.

All the preprocessing steps were executed using a combination of Python and MATLAB, BBCI Toolbox \cite{krepki2007berlin}, and EEGLAB \cite{Delorme2004EEGLAB}. Regarding the voice data, resampling was performed, adjusting the voice signals to a 22,050 Hz sample rate, while noise reduction was implemented using the noisereduce library~\cite{sainburg2020finding}.

\subsection{Feature Embeddings}
Spatial, temporal, and spectral information are recognized as vital components in speech-related brain signals, and vector-based brain embedding features have the capacity to convey the contextual meaning within these signals, as evidenced in prior studies. The embedding vector was generated using a combination of the common spatial pattern (CSP) to optimize spatial patterns and log-variance to extract temporal oscillation patterns. CSP, a technique for finding the most effective spatial filters using covariance matrices, plays a crucial role in deciphering brain signals associated with speech.

To minimize the disparities between the data distribution of spoken EEG and imagined EEG, we shared the CSP filters between both EEG signals \cite{lee2023AAAI}. Notably, the CSP filters were trained using imagined EEG, which exclusively contains pure brain signals, as opposed to spoken EEG, which may contain some level of noise. By sharing these CSP filters, the domain of spoken EEG was effectively adapted to the subspace of imagined EEG.


\subsection{Spatio-temporal Analysis}
To compare the brain activation while performing spoken speech and imagined speech, temporal, spatial, and spectral features were analyzed. For the spatial features, de-synchronization in the central lobe and synchronization in the temporal lobe was prominent. As for the spectral features, the high-frequency range above 90 Hz was dominantly synchronized. Both the imagined speech and the spoken speech have shown similar aspects, mainly desynchronization patterns around 500-1,250 ms dominantly in the high-frequency range, which supports that the BTS framework aims to adapt the two different domains of speech brain signal to each other \cite{lee2023AAAI}.

\section{RESULTS AND DISCUSSION}

\subsection{Feature Embeddings}
Fig.~\ref{embeding} depicts the time points of activation during spoken speech and imagined speech, displaying remarkably comparable results. The results demonstrate that the feature embedding of imagined speech follows the aspects of the spoken speech, which implies that the encoding architecture can capture the relevant time points from the imagined speech EEG, as well as the spoken speech EEG. Since this indicates that the CSP patterns and log-variances can represent several features of spoken or imagined voices, we used these embedding vectors as the input of the model to reconstruct user's voice.

\subsection{Embedding Vector Distributions}
While the distribution of spoken features and imagined features shows that they placed separately before sharing CSP patterns, features adapted using shared CSP patterns are placed in relatively closer latent space where classifier can be shared or domain shift can be performed in a narrow range (Fig.~\ref{tsnel}). Since our domain is mainly brain signal, which is very unclear compared to the image or speech domain, the clusters of each class show relatively soft margins.

\subsection{Spatio-spectral Features}
\begin{figure}[t]
\centering
    \includegraphics[width=\columnwidth]{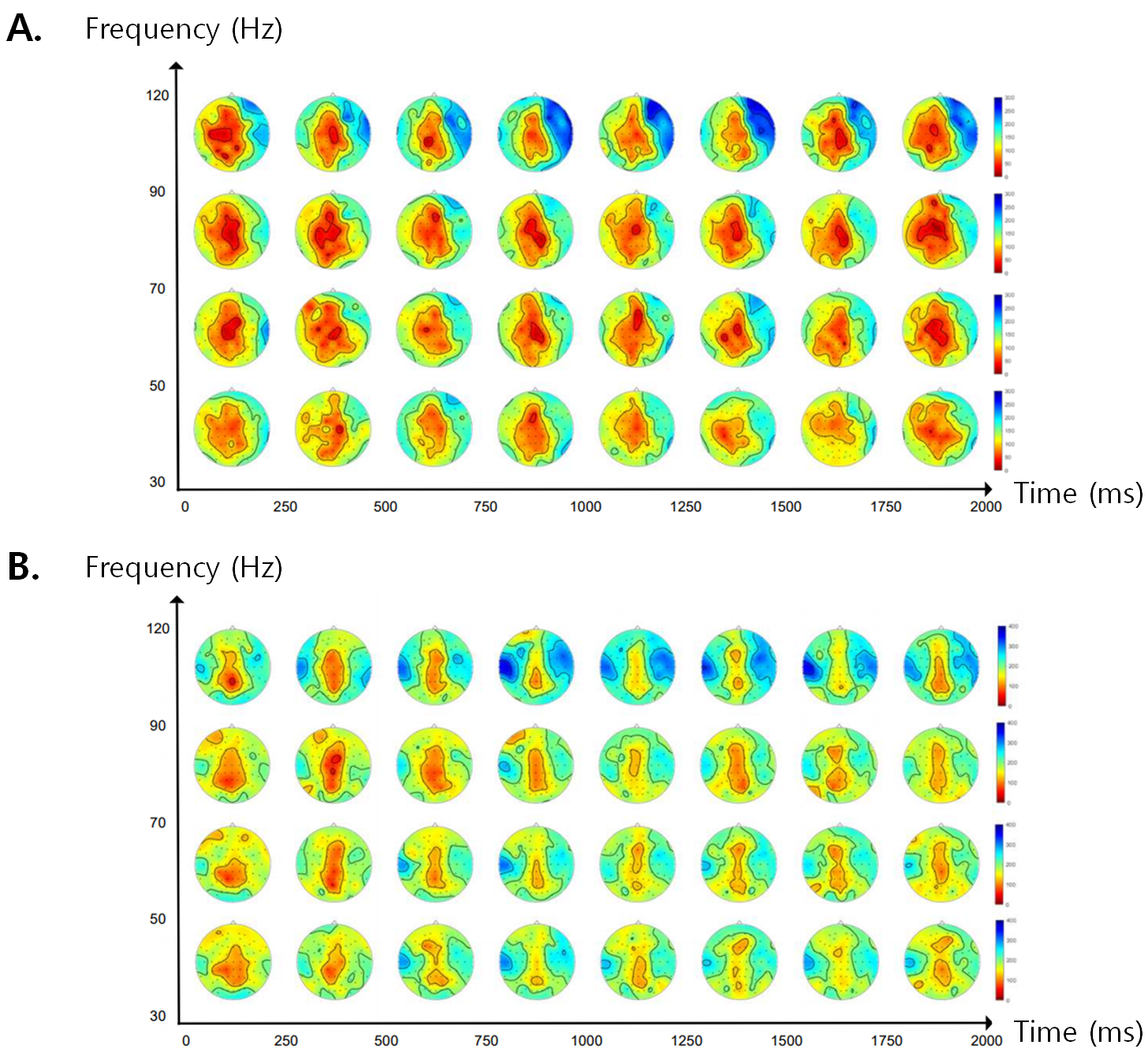}
    \caption{Temporal-spatio-spectral analysis of imagined speech 'thank you'. Changes of the power spectrum for (A) imagined speech and (B) spoken EEG is plotted for every 20$Hz$ frequency intervals in time shifts of 250$ms$.}
    \label{fig3}
\end{figure}

To explore disparities in brain activity during speech, we conducted an investigation into spatial and spectral characteristics, as depicted in Fig.~\ref{fig3}. Our analysis of spatial features unveiled noteworthy synchronization in the central lobe and desynchronization in the temporal lobe, in line with findings reported by Lee et al. \cite{Lee2019comparative}. In terms of spectral features, we identified strong synchronization in the high-gamma frequency range, specifically within the 30 to 120 Hz interval, as substantiated previous studies \cite{lee2019towards, anumanchipalli2019speech}. This high-gamma frequency band was selected for further examination. Furthermore, we noted distinct spatial characteristics for each text, with each text displaying its unique brain activation pattern. These observations suggest that various facets of speech may engage separate neural networks, and the observed brain activation patterns may mirror the motor processes that underlie speech production \cite{meng2022evidence}.

\subsection{Limitations and Future Works}

In our study, we were constrained by a dataset comprising only thirteen words. While we made every effort to prepare the model to produce novel words by incorporating all phonemes from those words, the outcomes did not meet our expectations. Our forthcoming objectives consist of extending our investigation by utilizing a dataset that encompasses a more extensive range of phonemes within sentences. Recent breakthroughs, such as the use of functional magnetic resonance imaging to decode semantic thoughts, have demonstrated the potential for brain signals to facilitate communication across language boundaries, offering the prospect of direct translation of neural signals. We anticipate that this enlarged dataset will permit us to investigate the neural features underlying speech process more precisely ~\cite{bang2021spatio, angrick2021real, suk2014predicting}. Additionally, we aim to implement our suggested methodology to invasive measures since speech BCI has both patient-specific and general applications. For patients, invasive methods focus on decoding articulator movements, which include acoustic and motor information. This could potentially serve as a communication tool for individuals faced with speech impairments~\cite{kim2019subject}. 




\section{CONCLUSION}
In this paper, we investigated the neural speech embeddings for speech synthesis from brain signals. Neural representation learning is an important issue in the field of non-invasive BCI since learning the optimal feature from noisy non-invasive brain signals plays a crucial role in the overall performance of decoding and generating speech. While current technologies are mostly limited in generating preliminary performance of word-level speech from non-invasive brain signals of imagined speech, further investigations on the neural representation learning may facilitate the development of a robust BTS system. The BTS framework can help disabled people who can only speak out few sounds, or patients who may potentially lose the ability to speak in the future (such as amyotrophic lateral sclerosis patients). We hope that the BTS technology to be positively used to improve the life for many people.

\bibliographystyle{jabbrv_IEEEtran}
\bibliography{REFERENCE}

\end{document}